\documentclass[conference]{IEEEtran}

\usepackage{bm}
\usepackage{flushend}
\usepackage{amsmath,amsfonts}
\usepackage{amsthm}
\usepackage{algorithmic}
\usepackage{algorithm}
\usepackage{array}
\usepackage{amsfonts}
\usepackage{graphicx}
\usepackage{textcomp}
\usepackage{stfloats}
\usepackage{url}
\usepackage{verbatim}
\usepackage{caption}
\usepackage{subcaption}
\usepackage[style=ieee,backend=bibtex, ]{biblatex}

\usepackage[dvipsnames]{xcolor}
\usepackage{tikz}
\usepackage{glossaries}

\newcommand{\x}{\bm{x}} %
\newcommand{\soAlg}{\boldsymbol{\xi}}%
\newcommand{\rotMat}{\mathbf{R}}%
\newcommand{\point}{\mathbf{r}}%
\newcommand{\y}{\bm{y}} %
\newcommand{\X}{\mathcal{X}} %
\newcommand{\Yt}{\mathcal{Y}_{target}} %
\newcommand{\R}{\mathcal{R}} %
\newcommand{\control}{\bm{u}} %
\newcommand{\U}{\mathcal{U}} %
\newcommand{\flow}{\bm{F}} %
\newcommand{\M}{\mathcal{M}} %
\newcommand{\uncertState}{\boldsymbol{\theta}}
\newcommand{\vel}{\boldsymbol{v}}

\newtheoremstyle{bfnote}%
  {}{}
  {\itshape}{}
  {\bfseries}{.}
  { }{\thmname{#1}\thmnumber{ #2}\thmnote{ (#3)}}

\theoremstyle{bfnote}

\newtheorem{assumption}{Assumption}[section]
\newtheorem{lemma}{Lemma}[section]
\newtheorem{theorem}{Theorem}[section]
\newtheorem{proposition}{Proposition}[section]
\newtheorem{definition}{Definition}[section]

\newacronym{GRP}{GRP}{Guaranteed Reachability Problem}
\newacronym{RA}{RA}{Reachability Analysis}
\newacronym{OCP}{OCP}{Optimal Control Problem}
\newacronym{ODE}{ODE}{Ordinary Differential Equation}
\newacronym{RTC}{RTC}{Reachable Temporal Coverage}
\newacronym{RG-OCP}{RG-OCP}{Reachability-Guaranteed \gls{OCP}}
\newacronym{NLP}{NLP}{Non-Linear Program}
\newacronym{MEGB}{MEGB}{Minimum Enclosing Geodesic Ball}

\begin{document}

\title{Reachability-Guaranteed Optimal Control for the Interception of Dynamic Targets under Uncertainty
}

\author{\IEEEauthorblockN{Tommaso Faraci\IEEEauthorrefmark{1}}
\IEEEauthorblockA{\textit{German Aerospace Center (DLR)} \\
Wessling, Germany \\
tommaso.faraci@dlr.de}
\and
\IEEEauthorblockN{Roberto Lampariello}
\IEEEauthorblockA{\textit{German Aerospace Center (DLR)} \\
Wessling, Germany \\
roberto.lampariello@dlr.de}
}

\maketitle
\begingroup\renewcommand\thefootnote{\IEEEauthorrefmark{1}}
\footnotetext{Tommaso Faraci is a PhD student at the University of Trento.}
\endgroup
\begin{abstract}
    Intercepting dynamic objects in uncertain environments involves a significant unresolved challenge in modern robotic systems.
    Current control approaches rely solely on estimated information, and results lack guarantees of robustness and feasibility. 
    In this work, we introduce a novel method to tackle the interception of targets whose motion is affected by known and bounded uncertainty. 
    Our approach introduces new techniques of reachability analysis for rigid bodies, leveraged to guarantee feasibility of interception under uncertain conditions.
    We then propose a Reachability-Guaranteed Optimal Control Problem, ensuring robustness and guaranteed reachability to a target set of configurations. 
    We demonstrate the methodology in the case study of an interception maneuver of a tumbling target in space. 
\end{abstract}
\section{Introduction}
Intercepting dynamic uncooperative targets is a major challenge in the current robotic landscape. Most techniques rely on static scenes or assume perfect knowledge of the scene's development in time. Only under these assumptions the approaches deliver trajectories guaranteeing obstacle avoidance and feasibility \cite{Tordesillas_2022}. 
In real scenarios, however, it is often the case that other object's properties are either unknown or derive from measurements affected by uncertainty. In most cases, this information is obtained from estimation procedures which deliver uncertain values \cite{lampariello2021robust}. 
Robust motion planning with reachability guarantees with respect to a designated target set thus remains an open problem. 
Current methodologies are not able to robustly perform in real life applications which require achieving close proximity, grasping and interacting with moving targets in $3D$-space (e.g. spacecrafts \cite{lampariello2021robust} and free-flying robots \cite{albee2021robust}).\par 
In this work, we address the problem of trajectory optimization for a rigid body in $3D$-space to guarantee the approach to and interception of a dynamic target, the \gls{GRP}. 
The target's motion parameters are affected by a known and bounded uncertainty, which we leverage to conservatively enclose all of its reachable trajectories. 
We then propose a nonlinear trajectory optimization method leveraging this enclosure, to deliver optimal trajectories with robustness and reachability guarantees. \par
\begin{figure}[!t]
    \centering
    \includegraphics[width=0.6\columnwidth]{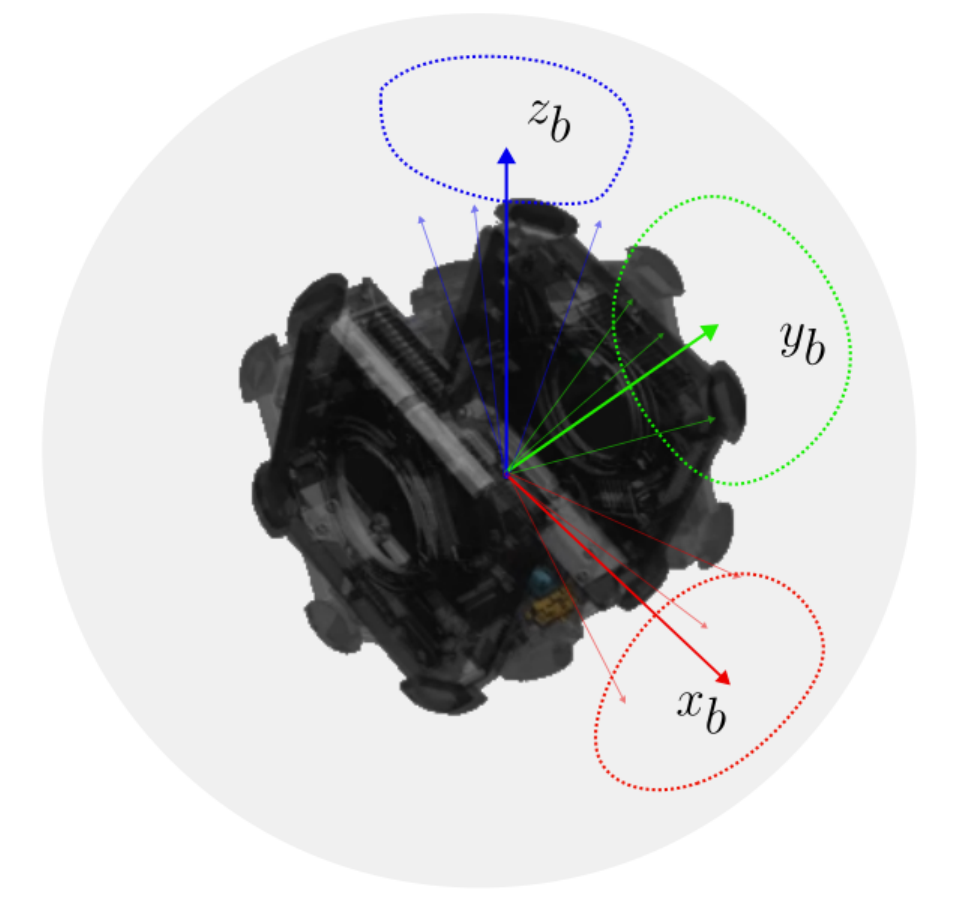}
    \caption{Visualization of reachable sets on $SO(3)$ for different realizations of simulated motion on NASA Astrobee robot \cite{smith2016astrobee}.
    All possible orientations are enclosed in a set, projected here on the unit-sphere for the robot's body-frame.}
    \label{fig:AstrobeeMultiRotation}
\end{figure}

{\bf Contributions:} We formalize the \gls{GRP}, which relies on the introduction of techniques of \gls{RA} on $SE(3)$. Furthermore, we propose a formulation of a \gls{RTC} to bound reachable trajectories in time. Finally, we introduce an \gls{OCP} to solve the \gls{GRP}.\par
{\bf Organization:} In Section \ref{section:RelatedWork} we introduce the current state-of-the-art methods for robust control and \gls{RA}. In Section \ref{section:NotationAndProblem} we introduce mathematical concepts and formulate the notion of \gls{GRP}. 
Section \ref{section:ConvexSetsSE3} introduces the notion of sets on $SE(3)$ and conservative convex enclosures on curved manifolds. 
Section \ref{section:TimeCoverage} describes derivation of the \gls{RTC}. 
In Section \ref{section:RG-OCP} we develop the \gls{RG-OCP}, which is applied in experiments in Section \ref{section:Experiments}. 
Finally, Section \ref{section:Conclusions} concludes this work by summarizing the results and introducing future work.

\section{Related Work}
\label{section:RelatedWork}
{\bf Motion planning under uncertainty} extends planning problems to account for \textit{aleatoric} (e.g., additive disturbances with no correlation in time) and \textit{epistemic} (e.g., parametric uncertainty on the system's parameters) uncertainty.
When probability distributions over uncertain quantities are available, \textit{chance-constrained} motion planning computes plans satisfying probabilistic constraints \cite{lathrop2021distributionally,lindemann2021robust}. 
However, in many applications, uncertainty is conservatively described as a bound rather than probability distributions \cite{abbasi2011improved, hillenbrand2005motion}, requiring to guarantee safe operation for all uncertainty realizations within these bounds. %
This task is referred to as \textit{robust motion planning}.
\\[1mm]
{\bf Robust motion planning} algorithms deliver plans while satisfying constraints for all possible uncertain quantities within known bounded sets. 
Often, methods rely on error bounds computed offline and enforced during online planning \cite{danielson2020robust,majumdar2016,singh2020robust}. 
Such methods are conservative and apply only to  stabilizable systems. 

Sample-based methodologies are proposed in~\cite{LewJansonEtAl2022, lew_estimating_2023,lew_convex_2024}, deriving error bounds on the reconstruction of reachable sets. These methods exploit properties of convex hulls of forward dynamic rollouts to constrain reachable sets at prescribed times. 
These methods are efficient and easily applicable to a variety of cases.
However, discrete forward propagation only delivers reachable sets at sampled times, introducing a lack of inter-sample guarantees between sets.
Moreover, tasks such as target interception \cite{ZhangTargetInterception, DongkyoungInterceptionRobotArm} and space debris removal \cite{mark2019review} require methods applicable on $SE(3)$. 
In this work, we propose a sampling-based temporal coverage for forward propagated dynamics, the \gls{RTC}, guaranteeing robustness by relying on general assumptions on the system's flow. 
The method allows to perform reachability analysis on $SE(3)$, enabling to actuate a \textit{guaranteed target approach}.
\\[1mm]
{\bf Guaranteed Target Approach} integrates techniques of \gls{RA} in a robust optimal control framework. 
Methods in \cite{lew_estimating_2023,lew_convex_2024} are used to robustly plan for systems affected by epistemic uncertainty, formulating constraint satisfaction on reachable sets leveraging properties of convex hulls on $\mathbb{R}^n$. 
However, the approach does not resolve the problem of guaranteeing reachability of all states in a target set.
We address the aforementioned limitations by tightly integrating sampling-based forward reachability analysis and an optimal control framework. 
The method guarantees the feasibility of all control trajectories, in order to ensure reachability of all the target's configurations.
To the best of the authors' knowledge, \gls{RA} has never been employed to robustly plan an approach maneuver guaranteeing full reachability with respect to a target. \par
First, we design a forward reachability method to estimate reachable sets of a dynamic target affected by aleatoric uncertainty on $SO(3)$.
Then, we expand the methods in \cite{lew_estimating_2023,lew_exact_2023} to factor robustness and reachability guarantees within an \gls{OCP} for a rigid body trajectory in $\mathbb{R}^n$.
We finally obtain the \gls{RG-OCP}, an efficient control methodology leveraging \gls{RA} to plan safely and robustly. We also argue that the method applies to a wide class of robotic systems.

\section{Background and Problem Definition}
\label{section:NotationAndProblem}
{\bf Differential geometry}. 
Let $\M\subseteq\mathbb{R}^n$ be a $k$-dimensional manifold without boundary. 
Equipped with the induced metric from the ambient Euclidean norm $\|\cdot\|$, $(\M,\|\cdot\|) $ is a Riemannian manifold \cite{do1992riemannian}. The geodesic distance on $\M$ between $p,q\in\M$ is denoted by 
$d^{\M}(p,q)$. Geodesics are curves $\gamma: I \subseteq \mathbb{R} \to \M$ satisfying the geodesic equation $\nabla_{\gamma '}\gamma ' = 0$ with respect to the Levi-Civita connection.
From \cite[Theorem~6.12]{gudmundsson2004introduction}, all geodesics $\gamma$ are locally minimizing with respect to the energy functional $E(\gamma) = \frac{1}{2} \int_a^b  \Vert \gamma ' \Vert^2$.
For any $p\in\M$, $T_p\M$ and $N_p\M$ denote the tangent and normal spaces of $\M$ \cite{Lee2012,do1992riemannian}. These are viewed as linear subspaces of $\mathbb{R}^n$.  
$(\M,\|\cdot\|) $ is obtained from its tangent space through the \textit{exponential map} $\exp() : E \subset T_p\M \to \M$ with $E$ defined in \cite{Lee2018}, $\exp_p(v) = \gamma_v(1)$. \par

{\bf Convexity on Riemanninan Manifolds} ties to the definition of a distance metric and geodesics. We report the following definition from Riemannian geometry:
\begin{definition}[Strong Convexity \cite{do1992riemannian}]
    A subset $S \subseteq \M$ is strongly convex if, for any two points $q_1, q_2$  in the closure $\bar{S}$ of $S$ there exists a unique minimizing geodesic $\gamma$ connecting $q_1$ and $q_2$ whose interior is contained in $S$.
\end{definition}
We further report a fundamental result from \cite{do1992riemannian}: 
\begin{proposition}[Convex Neighborhoods]\label{prop:ConvexNeighborhood}
    For any $p \in S$ there exist $\rho > 0 $ such that the geodesic ball $B(p, \rho) = \exp_{p}(B(0, \rho))$ is strongly convex. 
\end{proposition} 
The maximum value that $\rho$ can assume is the \textit{convexity radius}
\begin{equation}
    \rho_{conv}(p) = \max \{\rho\, \vert \, B(p,s)\,\text{strong conv.} \, \forall s \in [0,\rho] \}.
\end{equation}
The {\bf Guaranteed Reachability Problem} entails planning robust trajectories which guarantee reachability with respect to a set of states. 
We introduce dynamic systems under the following assumption: 
\begin{assumption}[System is Complete \& Bounded]{
    \label{ass:FlowComplete}
The continuous-time dynamical system }
\begin{equation}
    \dot{\x} = f (\boldsymbol{\phi})\label{eq:SystemDynamics}
\end{equation}
where $\boldsymbol{\phi} = [\x, \control, \uncertState] \in \Phi$ is an element of a smooth $n-$dimensional manifold $\Phi = \M_x \times\M_u \times \Theta$, has a smooth Lipschitz flow $\bm{F}^t (\phi): \Phi \to \M_x$ 
\begin{equation}
    \x (t_0) \equiv \x_0, \quad \bm{F}^t(\boldsymbol{\phi}) := \x_0 +  \int_{t_0}^{t_0 +t} f (\boldsymbol{\phi}(\tau)) d \tau \label{eq:DynFlow}
\end{equation}
is forward complete. Furthermore, $\Vert f (\boldsymbol{\phi}) \Vert$ is bounded.
\end{assumption}
We denote the reachable set of system (\ref{eq:SystemDynamics}) as $\Xi(t) = \flow^t(\Phi)$.\\
Assumption \ref{ass:FlowComplete} guarantees that a solution of the forward dynamics equation always exists, for a vector of states $\x$, inputs $\control$ and uncertain parameters $\uncertState$. 
The assumption above is always satisfied for continuos flows defined on compact manifolds.\par
In the course of this work, we focus on the following reachable sets: 
\begin{subequations}
    \label{eq:Systems}
    \begin{align}
        &\text{Controlled: }  &&\X(t) = \flow^t (\U) \label{eq:ControlledSys}\\ 
        &\text{Target: } &&\Yt (t) = \flow^t_{target}(\Theta) \label{eq:TargetSys}.
    \end{align}
\end{subequations}
Denoting how the controlled system's flow maps from the input $\control \in \U$, while the target's maps from uncertain parameters $\uncertState \in \Theta$.
\begin{definition}[{\bf \acrlong{GRP}}]
    Let (\ref{eq:Systems}) satisfy Assumption \ref{ass:FlowComplete}. 
    The target set $\Yt$ is reachable under guarantees from $\x_0$ at time $T_f$, if a set of admissible control trajectories $\U$ exists such that the reachable set $\X(T_f)$ satisfies: 
\end{definition}
\begin{equation}
    \X(T_f) \supset \Yt(T_f).
\end{equation}
\vspace{-5mm}
\section{Reachable Sets on \(SE(3)\)}
\label{section:ConvexSetsSE3}

Many applications in robotics involve $SE(3)$, the semidirect product  $SO(3) \times \mathbb{R}^3$. 
Hence, a subset  $\Phi \subset SE(3)$ can be expressed as $\Phi = S \times T$ with $S \in SO(3)$ and $T \in \mathbb{R}^3$.
To define uncertainty sets of poses, it is thus sufficient to compute sets of translations and rotations. 
As the theory of convexity in Euclidean space $\mathbb{R}^n$ is known, we focus our attention to define convex and compact sets on $SO(3)$.
Fig. \ref{fig:AstrobeeMultiRotation} allows to visualize reachable sets on $SO(3)$ for an Astrobee \cite{smith2016astrobee} subject to unstable motion. Uncertainty on the initial conditions is propagated as the robot tumbles.
\subsection{Strong Convexity on $SO(3)$} 
Proposition \ref{prop:ConvexNeighborhood} guarantees the existence of convex geodesic balls around any $\rotMat_o \in SO(3)$. 
The convexity radius on the manifold is known and bounded by the following Lemma: 
\begin{lemma}[Convexity Radius in $SO(3)$ \cite{hartley2010rotation}]
    \label{lemma:LimitingRadius}
    A closed ball $B(p, r) \subset SO(3)$ is \textit{strongly convex} iff $r < \frac{\pi}{2}$.
\end{lemma}
We can then formalize the concept of convex hull on $SO(3)$: 
\begin{definition}[Convex Hull on $SO(3)$\cite{hartley2010rotation}]\label{def:StronglyConvexHull}
    The strongly convex hull of set $S\subset SO(3)$ is the minimal strongly convex set that contains $S$. If it exists, then it is the intersection of all strongly convex sets containing $S$.
\end{definition}

In the following, we define strongly convex sets on $SO(3)$ exploiting the concept of a ball, which depends on the definition of a geodesic distance from a given orientation $\rotMat_o$. Moreover, as the Special Orthogonal group is a Lie group, we obtain a simple definition of the convex ball by expressing it in the Lie algebra $\mathfrak{so}(3)$.

Geodesics in \(SO(3)\) are expressed through the Lie exponential map, from $\rotMat_0$ to $\rotMat_1$ as $\gamma_{SO(3)}(t) = \rotMat_o \exp(t\log(\rotMat_0^T \rotMat_1))$, with $t \in [0,1]$. 
By defining a scalar product in the tangent space, we can define the notion of distance on $SO(3)$ $d^{SO(3)}(\rotMat_1, \rotMat_2) = \frac{1}{\sqrt{2}} \Vert \log(\rotMat_1^T \rotMat_2) \Vert_F$, corresponding to the minimal rotation between two orientations.
We leverage the Lie algebra $\soAlg^{\times}\in \mathfrak{so}(3)$ to define geodesic balls as in \cite{hartley2010rotation,gao_closure_2024}:
\begin{align}
    B(\rotMat_o, {\rho}) &= \left\{ \rotMat \in SO(3) \, \vert \, d^{SO(3)}(\rotMat_o, \rotMat) \leq \rho \right\} \nonumber \\
     &= \left\{\rotMat(\soAlg) \in SO(3) \, \vert \, \frac{1}{\sqrt{2}}\Vert \log(\rotMat_o^T \rotMat(\soAlg)) \Vert \leq \rho \right\} \nonumber 
\end{align}
introducing the Rodrigues formula, which corresponds to the exponential map of the Lie group 
\begin{equation}
    \rotMat(\soAlg) = \rotMat_o \exp(\soAlg^{\times}).
    \label{eq:SO3Map}
\end{equation}
We then define the geodesic ball as the image of a sphere in Euclidean space $S^2 \subset \mathbb{R}^3$: 
\begin{equation}
    B(\rotMat_o, {\rho})= \rotMat_o \exp(\soAlg^{\times})\label{eq:GeodesicBall}
\end{equation}
where $\soAlg \in B(0, \rho) = \left\{\soAlg \in \mathbb{R}^3 \, \vert \, \Vert \soAlg \Vert \leq \rho \right\} $.

\subsection{Reconstruction of Uncertainty Sets on $SO(3)$}
In the following, we propose a technique to reconstruct uncertainty sets of orientations from a finite number of samples. The method delivers a convex enclosure of the reachable set defining a geodesic ball the center of which is the mean orientation of all samples and the radius is the minimum distance enclosing them. 

Lemmas \ref{prop:ConvexNeighborhood} and \ref{lemma:LimitingRadius} guarantee the existence of a strongly convex neighborhood around $\rotMat_o$ as long as $\rho < \frac{\pi}{2}$.
Given a set of rotation samples $S_{\delta} \subset SO(3)$ where $S_{\delta} = \{\rotMat_i\}_{i=0}^{M_{\delta}-1}$ with $M_{\delta} \in \mathbb{R}^+$, we enclose its reconstruction $S$ conservatively through a \gls{MEGB} $B(\rotMat_o, \rho_{min})$ \cite{gao_closure_2024}, where: 
\begin{equation}
    \rotMat_{o} = \underset {\rotMat}{\text{arg min}} \sum_{i=0}^{N} d^{SO(3)}(\rotMat, \rotMat_i)^2 \label{eq:FrechetMean} 
\end{equation}
and 
\begin{equation}
    \rho_{min} = \max d(\rotMat_o, \rotMat_i).
\end{equation}
As long as $\rho_{min} <\frac{\pi}{2}$, the objective function (\ref{eq:FrechetMean}) is convex, thus the minimum exists and can be obtained \cite{hartley2010rotation}.

For $\Yt(t) = \flow_{Target}^t(\Theta)$, we introduce the $\delta$-cover \cite{Lee2012} $\Theta_{\delta} \subset \Theta$, which delivers, under suitable assumptions, the forward-rollouts $\mathcal{Y}_{\delta, target}(t)$. 
Therefore we can safely enclose $\mathcal{Y}_{\delta,\, target}(t)$ in a \gls{MEGB}.

\subsection{Reachable Convex sets for interception task}

To efficiently handle reachable sets on manifolds, we enclose them within convex hulls in Euclidean space. 
For a rigid body in $SO(3)$, each of its points $\point$ moves on the sphere $S^2$. The point moves on the spherical segment described by the geodesic ball obtained from: 
\begin{equation}
        \point_{\partial B} = \rotMat_o \exp(\soAlg^{\times}) \point 
\end{equation}
for $\soAlg \in \partial B(0, \rho) $. The resulting ball can be conservatively enclosed in any convex polygon arbitrarily defined, through a simple lift to $\mathbb{R}^3$.
See Section \ref{section:Experiments} for a visualization of the result used in this work.
\section{Temporal Coverage of Continuous Dynamics}
\label{section:TimeCoverage}

In \cite{lew_estimating_2023, lew_convex_2024} the authors do not account for the inter-sample flow development of their method concering the time discretization of sampled trajectories.
Lacking guarantees on the trajectory between time instants can lead to unsafe operations as constraints satisfaction cannot be verified. 
In the following, we define conservative time coverages for all possible trajectories, guaranteeing robustness and feasibility between time samples, completing the theory of reachable set reconstruction presented in \cite{lew_estimating_2023, lew_convex_2024}.

Under Assumption \ref{ass:FlowComplete}, it is guaranteed that any pair of $\x(t_j) \text{ and } \x(t_k)$ is bound to lie on the trajectory described by the flow (\ref{eq:DynFlow}). We can write
\begin{align}
    \bm{F}^{t_k}(\boldsymbol{\phi} ) - \bm{F}^{t_j}(\boldsymbol{\phi} ) = \int_{t_j}^{t_k} f (\boldsymbol{\phi} (\tau)) d \tau
\end{align}
which can be bounded upwards by introducing the following inequality: 
\begin{align}
    \int_{t_j}^{t_k} f (\boldsymbol{\phi} (\tau)) d \tau &\leq \int_{t_j}^{t_k} \Vert  f (\boldsymbol{\phi} (\tau)) \Vert d \tau \nonumber \\ 
    &\leq \int_{t_j}^{t_k} L_t d\tau   =  L_t \Vert t_k - t_j \Vert
\end{align}

and the flow is locally Lipschitz, with: 
\begin{equation}
    L_t = \max_{t_j \leq \tau \leq t_k } \Vert  f (\boldsymbol{\phi}(\tau)) \Vert 
\end{equation}
where the operator $\Vert \cdot \Vert$ denotes the Euclidean matrix norm.
\begin{figure*}[!t]
        \centering
        \includegraphics[width=0.8\textwidth]{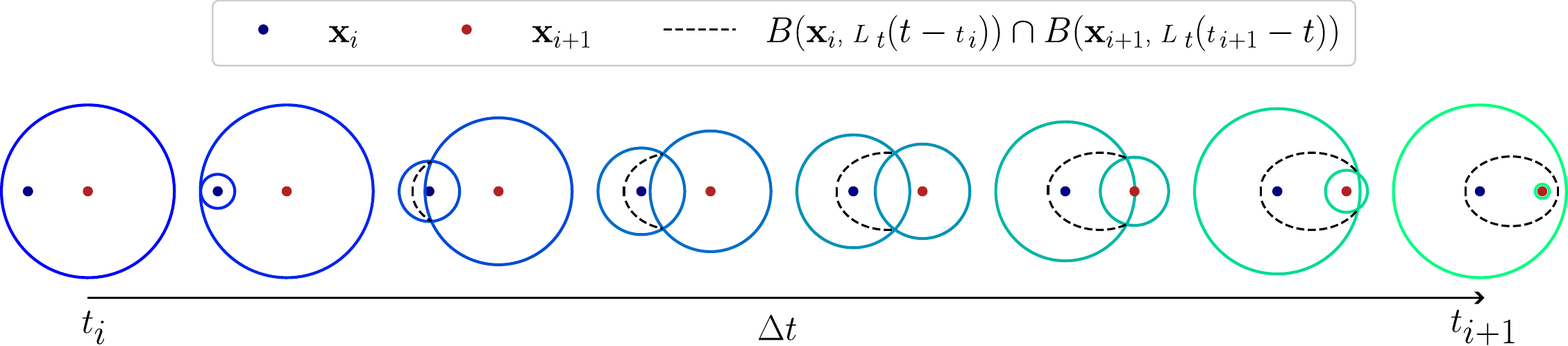}
        \caption{For each $t \in [t_i, t_{i+1}]$, $\flow^t(\boldsymbol{\phi})$ is bounded in time by $L_t$. Between states $\x_i$ and $\x_{i+1}$, the locus of the intersection of the two balls defines the region bounding all possible trajectories.}
        \label{fig_2}
\end{figure*}
Letting $\x_i = \x(t_i) $ and $\x_{i+1} = \x(t_{i+1})$, each element on the trajectory $\x(t)$ with $t_i \leq t \leq t_{i+1}$ must lie in the set $\mathcal{E}$: 
\begin{equation}
    \mathcal{E} =    B(\x_i ,  L_t (t - t_i)) \cap B(\x_{i+1} , L_t (t_{i+1} - t) )
    \label{eq:TimeBoundEllipsoid}
\end{equation}
Each point  $\x_{\partial \mathcal{E}} \in \partial \mathcal{E}$ satisfies the following condition: 
\begin{align}
    \begin{cases}
        d^{\mathcal{M}}(\x_{i}, \x_{\partial \mathcal{E}}) = L_t (t - t_i) \\ 
        d^{\mathcal{M}}(\x_{\partial \mathcal{E}}, \x_{i+1}) = L_t (t_{i+1} - t)
    \end{cases}
\end{align}
which, summing the two equations together, delivers: 
\begin{equation}
    d^{\mathcal{M}}(\x_{i}, \x_{\partial \mathcal{E}}) + d^{\mathcal{M}}(\x_{\partial \mathcal{E}}, \x_{i+1}) = L_t \Delta t. 
    \label{eq:LocusEllipse}
\end{equation}
Eq. (\ref{eq:LocusEllipse}) represents an ellipse with semi-major axis $\frac{L_t \,\Delta t}{2}$ on $\mathcal{M}_x$. 
A visual representation of how the bounding ellipse can be obtained is depicted in Fig. \ref{fig_2}.

The ellipse is aligned with the geodesic connecting the two focii $\x_i, \, \x_{i+1}$.
In $\mathbb{R}^n$, the ellipse can be represented as a quadratic form: 
\begin{equation}
    (\x - \x_c)^T Q (\x - \x_c)   = 1 \label{eq:EllipsoidEquation}
\end{equation}
where $\x_{c} = \frac{\x_{i} + \x_{i+1}}{2}$. 

In the context of \gls{RA}, in accordance with the theoretical results in \cite{lew_convex_2024}, the reachable set  $\Xi(t)$, of (\ref{eq:SystemDynamics}), is reconstructed from sampled dynamic rollouts $\Xi_{\delta} (t)= \flow^t(\Phi_{\delta})$. The reconstruction's Hausdorff distance to the real set $\Xi(t)$ is at most $\varepsilon$. 
Therefore, we propose an enclosure for the reachable trajectories of a neighborhood of size $\varepsilon$ around two solutions of the flow $\x_i \in \Xi_{\delta}(t_i), \x_{i+1} \in \Xi_{\delta}(t_{i+1})$ connected by a trajectory $\x(t)$ in time, by defining: 
\begin{equation}
    \mathcal{T}_{i,k} = \mathcal{E}(\x_{i,k}, \x_{i+1,k}) \oplus B(0, \varepsilon) \label{eq:paddedEllipsoid}
\end{equation}
where we have used the $\oplus$ to define the Minkowski addition of the ellipsoid $\mathcal{E}$ and the ball $B$.
This formulation allows us to define the set of all reachable trajectories connecting each state by defining the \gls{RTC}. 
\begin{definition}[\acrfull{RTC}]
    \label{def:RTC}
    Let system (\ref{eq:SystemDynamics}) satisfy Ass. \ref{ass:FlowComplete}. Let $ \Xi_{\delta, \,i} = \flow^{t_i}(\Phi_{\delta})$ and $ \Xi_{\delta, \,i+1} = \flow^{t_{i+1}}(\Phi_{\delta})$ be its forward rollouts obtained at times $t_i$ and $t_{i+1}$ from $\delta$-cover $\Phi_{\delta} \subset \Phi$.
    The \gls{RTC} of the reachable set $\Xi(t)$ with $t_i \leq t \leq t_{i+1}$ is defined as: 
    \begin{equation}
        \mathcal{R}_i = H( \bigcup_{k=0}^M \mathcal{T}_{i,k}),
        \label{eq:ReachableTimeCovering_i}
    \end{equation}
    where $H(\cdot)$ denotes the convex hull operation.
\end{definition}

\begin{figure}[!h]
\centering
\includegraphics[width=0.8\columnwidth]{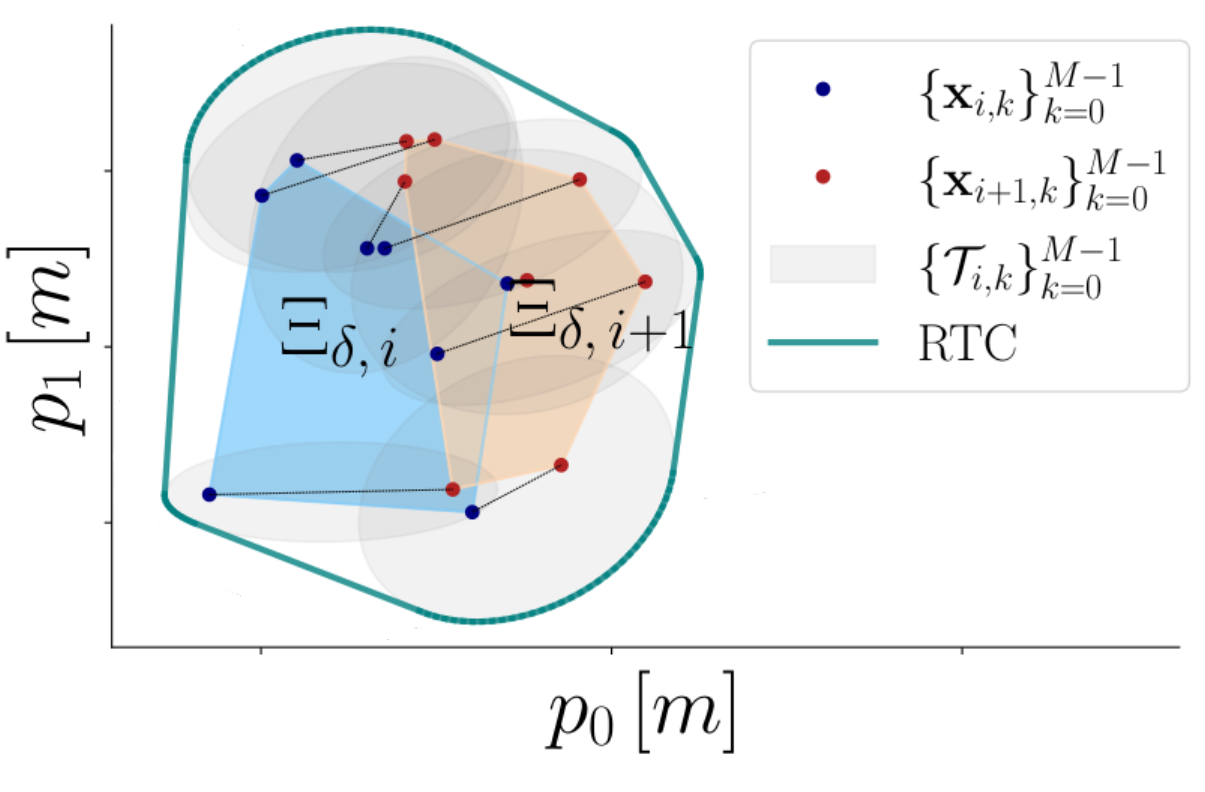}
\caption{Sketch of time coverages of the sets $\Xi_i$ (\protect\tikz[baseline]{\protect\fill[CornflowerBlue!70!white] (0,0) rectangle (0.2,0.2);})  and $\Xi_{i+1}$ (\protect\tikz[baseline]{\protect\fill[RedOrange!40!white] (0,0) rectangle (0.2,0.2);}) by union of bounding ellipsoids $\mathcal{T}$ (\protect\tikz[baseline]{\protect\fill[Gray!40!white] (0,0) rectangle (0.2,0.2);}) to obtain the boundary of the \gls{RTC} $\partial \R_i$ (\protect\tikz[baseline]{\protect\draw[blue!50!green] (0,0.05) -- (0.5,0.05);})}
\label{fig:CompleteTimeEnclosure}
\end{figure}

For ease of understanding, Fig. \ref{fig:CompleteTimeEnclosure} depicts the \gls{RTC} between two sets in $2-$dimensional space.

We now introduce the most relevant theoretical result of this work, which states that \glspl{RTC} between two consecutive reachable sets conservatively bound all reachable trajectories connecting them:

\begin{theorem}[Reachable time-covering contains Reachable sets]
    \label{Theo1}
    The \gls{RTC} $ \mathcal{R}_i$  strictly contains all reachable sets $\Xi(t)$ for $ t_i \leq t \leq t_{i+1}$: 
    \begin{equation}
        \mathcal{R}_i  \supset \bigcup_{t=t_i}^{t_{i+1}}\Xi(t)
    \end{equation}
    \vspace{-5mm}
\end{theorem}
The proof can be found in the Appendix.
Theorem \ref{Theo1} guarantees that we can conservatively bound all flows over time, by considering the union of all \glspl{RTC} over $N$ time-steps:
\begin{equation}
\mathcal{R} = \bigcup_{i}^N \mathcal{R}_i .
\end{equation}
\vspace{-5mm}
\section{Reachability-guaranteed Optimal Control}\label{section:RG-OCP}
\vspace{-1mm}
We present the main contribution of this work. 
We formulate the \gls{RG-OCP} in order to solve the \gls{GRP}, building on the theoretical results obtained in Sections \ref{section:ConvexSetsSE3}-\ref{section:TimeCoverage} as follows.
\begin{definition}[\acrlong{RG-OCP}]\label{def:RGOCP}
    Let systems (\ref{eq:Systems}) satisfy Assumptions \ref{ass:FlowComplete}. 
    The solution of the \gls{RG-OCP} is the sequence of control sets $\U^{*}(t)$ obtained from \gls{OCP}:
    \begin{subequations}
    \begin{align}
        \U^{*}(t) = \underset {\U(t)}{\text{arg min}}&\int_0^{T_t} L(\X(t), \U(t)) dt \label{RGOCP:Hamiltonian}\\
        \text{subject to} \nonumber\\
        \U(t) &\subseteq \mathbf{U}_{adm}\, \forall t \in [0, \, T_f] \label{RGOCP:Constr:Control}\\
        \X(t)  &\subseteq \mathbf{X}_{adm}\, \forall t \in [0, \, T_f] \label{RGOCP:Constr:State} \\
        \dot{\X}(t)  &\subseteq \dot{\mathbf{X}}_{adm} \, \forall t \in [0, \, T_f]\label{RGOCP:Constr:StateDot} \\
        \mathcal{O} \cap \R &= \emptyset \label{RGOCP:Constr:Obstacle}\\
        \Yt& \subseteq  \X(T_f) \label{RGOCP:Constr:Target} \\ 
        \x_{nom}(T_f)&  \in \Yt \label{RGOCP:Constr:Endpoint}
    \end{align}
    \end{subequations}
\end{definition}
where (\ref{RGOCP:Hamiltonian}) describes the cost function applied to the state and control sets.
Eq. (\ref{RGOCP:Constr:Control}) is the set-valued control constraint, with box-constraint $\mathbf{U}_{adm}$ representing the set of admissible input values.
Correspondingly, (\ref{RGOCP:Constr:State})-(\ref{RGOCP:Constr:StateDot}) are the state and state derivatives constraints applied to the entirety of the reachable set by defining box-bounds $\mathbf{X}_{adm}, \, \dot{\mathbf{X}}_{adm} $. 
Eq. (\ref{RGOCP:Constr:Obstacle}) ensures that the \gls{RTC} has no intersection with any obstacle in the configuration space.
(\ref{RGOCP:Constr:Target})-(\ref{RGOCP:Constr:Endpoint}) enforces guaranteed reachability, introducing nominal trajectory $\x_{nom}$ steering the system within the target set. 

In the following, we present an approach for the resolution of the \gls{RG-OCP} by exploiting the techniques of \gls{RA} defined by \cite{lew_exact_2023,lew_convex_2024} and employing a sampling-technique to write an efficient \gls{NLP}.
\vspace{-1mm}
\subsection{Guaranteed Target Reachability}\label{sec:RG-OCP/GuaranteedReach}
In the context of this section, we aim at defining a condition that allows to provably guarantee that a set of states is reachable within a time horizon $T_f$ by system (\ref{eq:ControlledSys}) under Assumption \ref{ass:FlowComplete}. 
Given initial conditions $\x_0$, this requires the existence of a set of admissible control sequences $\U(t)$ steering system (\ref{eq:SystemDynamics}) from $\x_0$ to all $\y_{target} \in \Yt$. \\
{\bf Controlled system \gls{RA} through convexification} allows to efficiently construct $\U(t)$, leveraging the theoretical guarantees obtained by \cite{lew_estimating_2023,lew_convex_2024} for \gls{RA}. 
In particular, we can write the reachable set $\X(t)$ as the set of all flows under the control sequences $\U(t)$, $\X(t) = \flow^t(\U(t))$.
If $\flow^t : \U(t) \subseteq \mathbb{R}^m \rightarrow \X(t) \subseteq \mathbb{R}^n $ is  $\mathcal{C}^1$, writing $\U(t)$ as an $r$-smooth convex set allows us to leverage the results from \cite{lew_convex_2024}.
It is thus sufficient to forward propagate a $\delta$-cover $\U_{\delta}(t) \subseteq \U(t)$, to reconstruct reachable sets from $\X_{\delta}(t)$.
Thus, we conservatively define $\U(t)$ as a ball of constant radius $R_{\delta}$ around a nominal control trajectory $\control^*(t)$: 
\begin{equation}
    \U(t) = \control^* \oplus B(0, R_{\delta}) \label{eq:ControlRadiusSet}.
\end{equation}

\begin{figure}[!h]
    \centering
    \includegraphics[width=0.5\columnwidth]{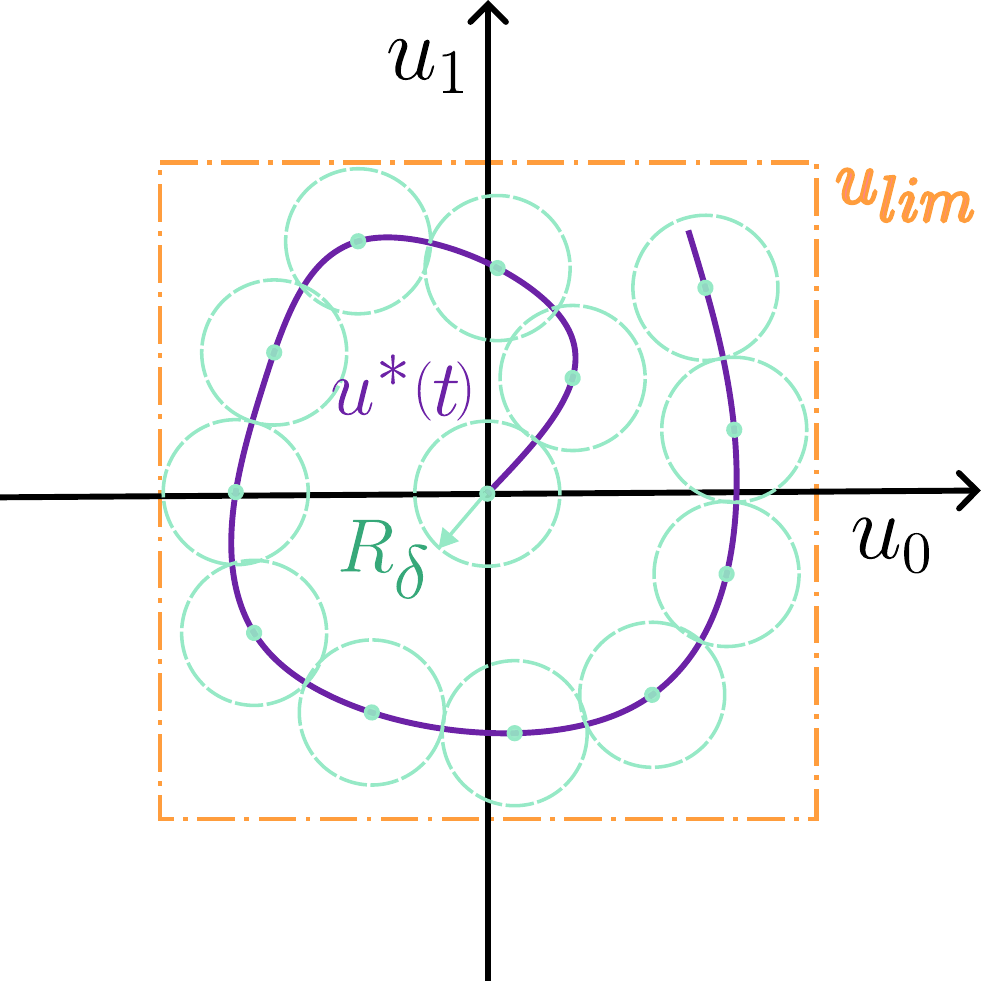}
    \caption{Equation (\ref{eq:ControlRadiusSet}) states that the set of admissible inputs $\U(t)$ is conservatively contained by a ball of radius $R_{\delta}$ around the optimal input trajectory $\control^*(t)$.}
    \label{fig:InputUncertainty}
\end{figure}

We can efficiently define a $\delta$-cover $ \U_{\delta}(t) $ of $\U(t)$ as $ \U_{\delta}(t) = \control^*(t) + \{R_{\delta}\mathbf{s}_k\}_{k=1}^{M}$, using samples $\{R_{\delta}\mathbf{s}_k\}_{k=1}^{M} \subset \partial B(0, R_{\delta})$, as depicted in Fig. \ref{fig:InputUncertainty}.

The expression in (\ref{eq:ControlRadiusSet}) allows to easily formulate the control constraint (\ref{RGOCP:Constr:Control}) as follows:
\begin{equation}
    -u_{lim} \leq \Vert \control^*(t) + R_{\delta}\Vert_{\infty} \leq u_{lim}.
\end{equation}
The triangle inequality guarantees $\Vert \control^*(t) + R_{\delta}\Vert_{\infty} \leq \Vert \control^*(t) \Vert_{\infty}  + \Vert R_{\delta} \Vert_{\infty} = \Vert \control^*(t) \Vert_{\infty}  +R_{\delta} $. 
Thus, we write the constraint as a tractable finite-dimensional relaxation for $n_{\control}$-dimensional control vectors on $N$ control segments: 
\begin{align}
    -u_{lim}  \leq u_i (t_j) + R_{\delta}\leq u_{lim} \quad &\text{for }i=0, \dots, n_u-1, \label{eq:ControlConstrRelaxation} \\ &\text{and }j=0, \dots, N-1 \nonumber. 
\end{align}

Performing forward propagation of $\U_{\delta}$, we obtain from \cite{lew_estimating_2023}:
\begin{equation}
    \X(t_i) \subseteq H(\X_{\delta}(t_i)) \oplus B(0, \epsilon). 
\end{equation}
we can easily formulate box constraints (\ref{RGOCP:Constr:State})-(\ref{RGOCP:Constr:StateDot}) as finite-dimensional constraints:
\begin{align}
    \x_{min} + \epsilon &\leq \x_{k}(t_j) \leq \x_{max} - \epsilon \, &\text{for } k = 0, \dots, M-1 \label{eq:StateConstrRelaxation}\\ 
    -\vel_{sup}  &\leq \dot{\x}_{k}(t_j) \leq \vel_{sup}   \, &\text{and } j = 0, \dots, N-1.
\end{align}

{\bf Set-containment between sets reconstructed from samples} allows to guarantee reachability of $\Yt(T_f)$ expressed as a convex polygon $\mathcal{P}^{n_{x}}$ with $M_{target}$ vertices:
\begin{equation}
    \Yt(T_f) \subseteq \mathcal{P}^{n_x} =  H(\{\y_{target, k}\}_{k=0}^{M_{target}-1}). \label{eq:TargetSetAsPoints}
\end{equation}
The convex hull operation $H(\cdot)$ in $\mathbb{R}^{n_x}$ performed in (\ref{eq:TargetSetAsPoints}) exploits convex combinations in the target set to guarantee containment.
\begin{lemma}[Convex combination leads to set containment]
    \label{Lemma2}
    Let sets $A,B\, \subset \mathbb{R}^n$, with $ A = \{x_0, \dots, x_{h-1}\}$ and $B = \{y_0, \dots, y_{g-1}\}$. 
    If, for each $\bar{y} \in \partial H(B)$ there exists coefficients $\lambda_i$ such that 
\end{lemma}
    \begin{equation}
        \bar{y} = \sum_{i=0}^{h-1} \lambda_i x_i 
    \end{equation}
    with $\sum_{i=0}^{h-1} \lambda_i = 1$ and  $\lambda_i \geq 0$, then $B \subset H(A)$.

We rewrite the constraint as: 
\begin{equation}
    \y_{target, \, i} = \sum_{j=0}^{M-1} \lambda_{i,j} \x_{j}\, \text{for } i = 0, \dots, M_{target}-1
\end{equation}
in matrix form:
\begin{equation}
    \mathbf{Y}_{target} - \mathbf{\Lambda} \mathbf{X}_{\delta}(T_f) = 0
\end{equation}
where $\mathbf{Y}_{target} = [\y_0^T, \dots, \y_{M_{target}-1}^T]^T$, $\mathbf{X}_{\delta}(T_f) = [\x_0^T(T_f), \dots, \x_{M-1}^T(T_f)]^T$ and $\mathbf{\Lambda}$ is the matrix of convex coefficients $\lambda_{i,j}$.

Finally, the nominal trajectory $\x_{nom}(t)$ is constrained to reach a nominal endpoint within the target set $\y_{nom}(T_f)$ up to a tolerance $\delta_{nom}$. 
The nominal endpoint is obtained by following the nominal trajectory of the target $\y_{nom}(T_f)$.
\vspace{-2mm}
\subsection{Time enclosures}
We constrain all trajectories between consecutive sets $\X_{\delta}(t_i)$ and $\X_{\delta}(t_{i+1})$ by conservatively applying constraints on the ellipsoids generating the \gls{RTC}.
Describing constraints (\ref{RGOCP:Constr:State}) to (\ref{RGOCP:Constr:Obstacle}) as convex decompositions (e.g. \cite{deits2015computing, liu2017planning}), we exploit the convex hull formulation of the \gls{RTC} to enforce them.
In $\mathbb{R}^{n_x}$, if the constraints are formulated as the intersection of hyperplanes describing obstacle-free space $\mathcal{C}_{free}$ with $n_{O}$ constraints:
\begin{equation}
    \mathcal{C}_{j} = \{\x \in \mathbb{R}^n: \mathbf{p}_j^T \x + h_j \leq 0, \, h_j \in \mathbb{R}, \Vert \mathbf{p}_j \Vert = 1 \}
\end{equation}

we enforce the constraint as in \cite{lew2020samplingbased}, by specifying for each ellipsoid $\{\{\mathcal{E}_{i,k}\}_{k=0}^{M-1}\}_{i=0}^{N-1}$: 
\begin{equation}
    \mathbf{p}_j^T \x_{c,\,i,k} + \sqrt{(\mathbf{p}_j^T Q_{i,k} \mathbf{p}_j)} \leq - h_j \,
    \text{for } j = 0, \dots, n_{\mathcal{O}} -1.
\end{equation}
For the sake of clarity, we summarize this nonlinear constraint as $g(\mathcal{E}, \mathcal{O}) \leq 0$.
\subsection{Controlled Lipschitz Constants}
The controlled system $f(\U^*(t))$ obtained by solving the \gls{RG-OCP}, by definition, has bounded velocity at discrete times $\Vert \mathbf{v}_{k}(t_i) \Vert_{\infty} \leq \mathbf{v}_{sup}$. 
Hence, the flow is bounded through the maximal achievable acceleration $\mathbf{a}_{sup}$: 
\begin{align}
    L_t = \Vert \vel_{sup} \Vert_2 +  \frac{\Delta t}{2}  \Vert \mathbf{a}_{sup} \Vert_2 = \vel_{lim}.
\end{align}

In general, we assume it is possible to express the maximal acceleration as $\mathbf{a}_{sup} = \sup(f(\x, \control)) = f(\x_{lim}, \control_{lim})$. 
Assuming this quantity is known, the Lipschitz constant of the flow can be artificially fixed defining: 
\begin{equation}
    \vel_{sup} = \vel_{lim} - \frac{\Delta t}{2} \mathbf{a}_{sup}
\end{equation}
This results guarantees that the flow's velocity never exceeds the value of $\vel_{lim}$ along a trajectory.

\subsection{Finite-dimensional Relaxation of \gls{RG-OCP}}
Under the relaxations introduced in the previous paragraphs, we define a numerically efficient formulation of the \gls{RG-OCP}. 
\begin{definition}[Finite-dimensional \gls{RG-OCP}]\label{def:FiniteRG-OCP}
    Let systems (\ref{eq:Systems}) satisfy Assumptions \ref{ass:FlowComplete}. 
    The solution of the Finite-dimensional Relaxation of \gls{RG-OCP} is the sequence of control sets defined by the tuple $(\mathbf{U}^*, R_{\delta})$, with $\mathbf{U} = [\control_0, \dots , \control_{N-1}]$. 
    \begin{subequations}
    \begin{align}
        \mathbf{U}^*, R_{\delta} =& \underset {\mathbf{U}, R_{\delta}}{\text{arg min}}\sum_{i=0}^{N-1}l(\mathbf{X}_i, \control_i) \label{RGOCP:Finite:Hamiltonian}\\
        \text{subject to } &\nonumber\\
        \bigwedge_{i=0}^{N-1}&\control_i + R_{\delta} \in \mathbf{U}_{adm} \\
        \bigwedge_{i=0}^{N-1}& \X_i \in \mathbf{X}_{adm}  \\
        \bigwedge_{i=0}^{N-1}& \dot{\X_i} \in \dot{\mathbf{X}}_{adm} , \\
        g(\mathcal{E}&, \mathcal{O}) \leq 0 \\
        \mathbf{Y}_{target}& = \mathbf{\Lambda} \mathbf{X}_{\delta}(T_f) \\
        \x_{nom}& \in B(\y_{nom}, \delta_{nom}) .
    \end{align}
    \end{subequations}
\end{definition}
We argue that, if a solution to Def. \ref{def:FiniteRG-OCP} is found, then it conservatively encloses the control trajectories $\U^*(t)$, solutions of Def. \ref{def:RGOCP}.

\section{Experiments}\label{section:Experiments}
In the following, we demonstrate the results of the \gls{RG-OCP} by numerically resolving it on the case of a spacecraft approaching a tumbling target for grasping. 
All methods are applied on a laptop, equipped with a \texttt{Intel i5-4300U} with $1.90 GHz$ and $8$ Gb of \texttt{RAM}.

\begin{figure}[!h]
    \centering
    \includegraphics[width=0.7\columnwidth]{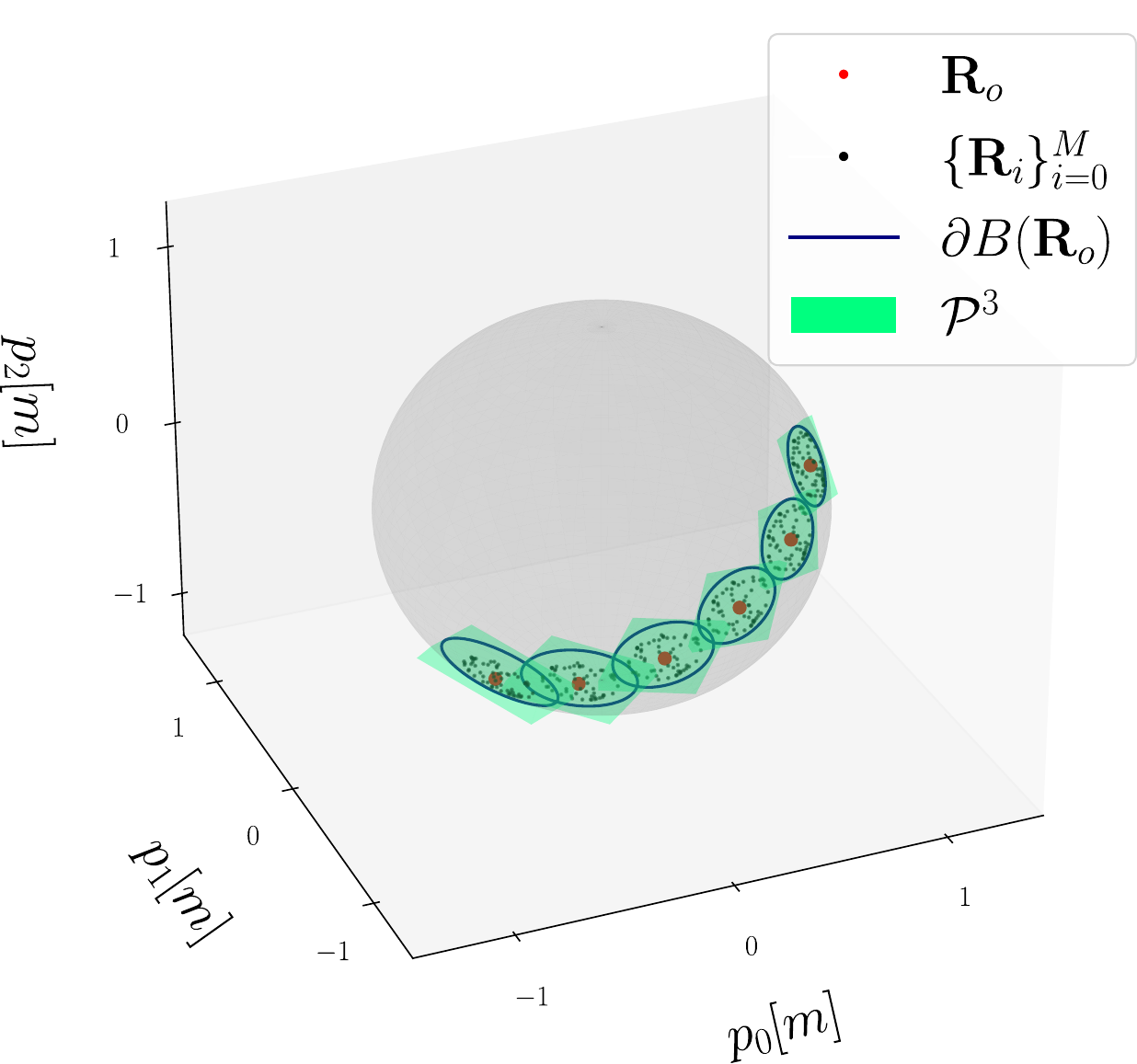}
    \caption{Reachable sets on $SO(3)$ projected onto the unit sphere and lifted to $\mathbb{R}^3$ to construct a convex polygon (\protect\tikz[baseline]{\protect\fill[green!70!white] (0,0) rectangle (0.2,0.2);}). The last set on the lower left is the target of the guaranteed approach.} 
    \label{fig:SO3Reach}
\end{figure}

\subsection{Guaranteed Approach for Interception of Unknown Tumbling Target}

We study the case of guaranteeing the feasibility of an interception task for a $3D$ freely rotating rigid body, described by the rotational dynamics:
\begin{subequations}
    \begin{align}
        \dot{\rotMat} &= \boldsymbol{\omega}^{\times} \rotMat \\
        \dot{\boldsymbol{\omega}} &= \mathbf{J}^{-1}(-\boldsymbol{\omega}\times \mathbf{J} \boldsymbol{\omega} ).
        \vspace{-2mm}
    \end{align}
\end{subequations}
The target's inertia in the target body frame is $\mathbf{J} = diag(29.2, 30, 38.4)$ and the angular velocity is $\boldsymbol{\omega} = [0.0, 0.0698, 0.0]^T$. It is worth noting that this represents an unstable equilibrium.
The body's orientation uncertainty is expressed as parametric uncertainty according to (\ref{eq:GeodesicBall}), assuming $\rho_{SO(3)} = 0.17 rad = 10^{\circ}$ around $\rotMat_o = \mathbf{I}_{3}$.

{\bf \gls{RA} for a tumbling spacecraft} is performed on $SO(3)$ for a time horizon of $30 \,s$ with $M = 30$. 
We find the \gls{MEGB} on $SO(3)$ and then lift the result on $\mathbb{R}^3$ for a grasping point located at $\point_{grasp, B} = [1, 0, 0]^T$ in the body frame. 
Results are obtained in $7 \,ms$ and  are shown in Fig. \ref{fig:SO3Reach}.
{\bf Guaranteed approach} is achieved with a spacecraft of $m=32.0\,kg$, capturing a small satellite rotating around the point $\mathbf{c}_{target} = [5.5, 0, 0]$.
For this specific case, the cost function aims to minimze the control set $\U(t)$. Hence, we formulate the functional introduced in (\ref{RGOCP:Finite:Hamiltonian}) as: 
\begin{equation}
    \sum_{i=0}^{N-1}l(\mathbf{X}_i, \control_i) = \sum_{i=0}^{N-1} \frac{1}{2}  (\control_i ^T \control_i  + R_{\delta}^2), 
\end{equation}
which accounts for both the nominal trajectory and the dimension of the control set.

\begin{figure}[!h]
    \centering
    \includegraphics[width=0.9\columnwidth]{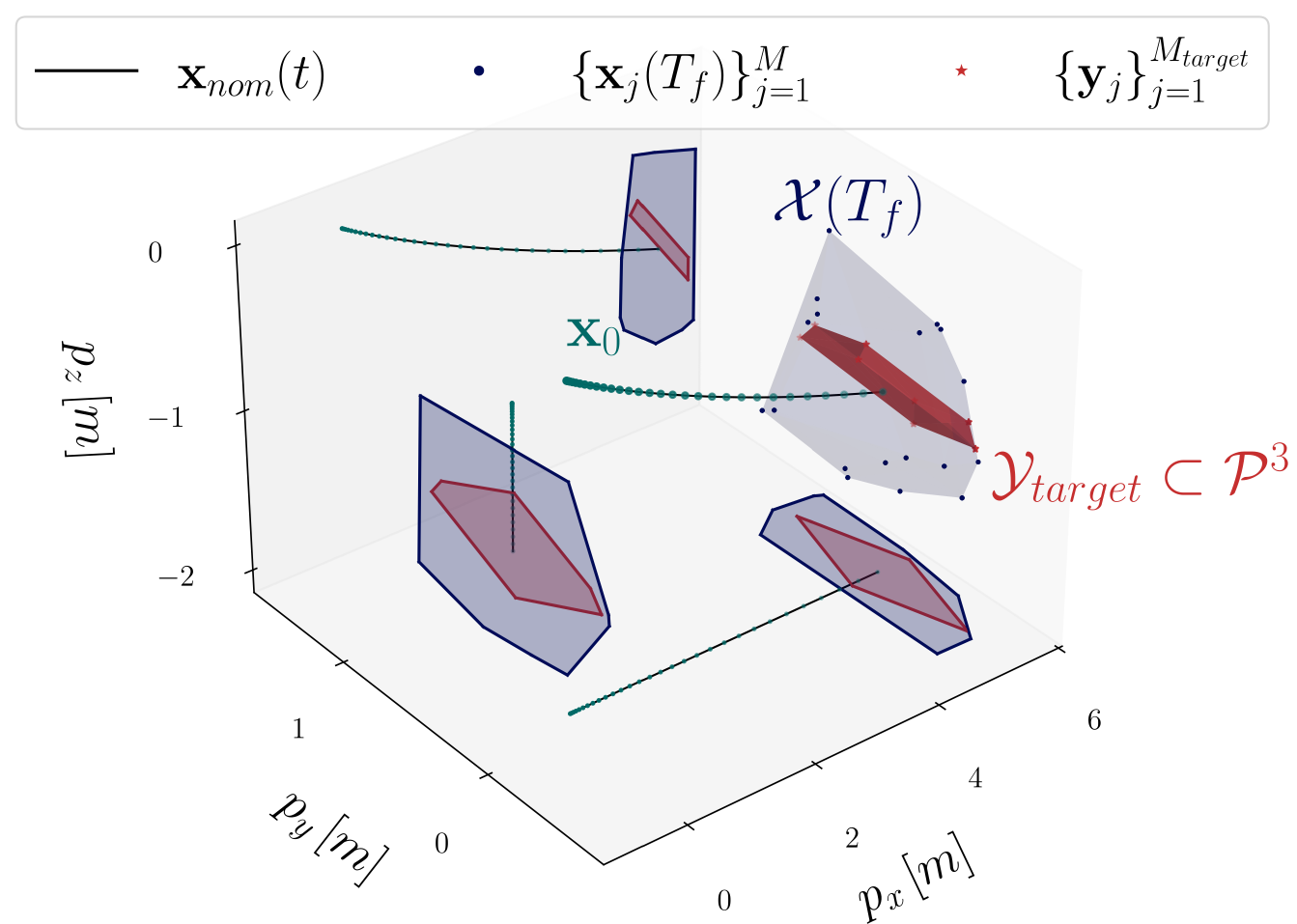}
    \caption{Result of \gls{RG-OCP} for interception of a tumbling target in $3D$ space, projected on the $3$ planes defined by the world frame axis.
    The controlled set (\protect\tikz[baseline]{\protect\fill[blue!70!white] (0,0) rectangle (0.2,0.2);}) tightly encloses the red polygon (\protect\tikz[baseline]{\protect\fill[red!70!white] (0,0) rectangle (0.2,0.2);}) which conservatively approximates the target set shown in Fig. \ref{fig:SO3Reach}, guaranteeing feasibility of the interception maneuver.}
    \label{fig:RG-OCPRes}
\end{figure}

The time horizon is restricted to $T_f = 30 \,s$. 
The solution is found in $2658 \,ms$ using $M_{target} = 8$ and $M = 32$ and a discretization of $\Delta t = 1 \,s$, thus $N=30$. 
The results are presented in Fig. \ref{fig:RG-OCPRes}, demonstrating the effectiveness of the method. The controlled system is able to perform all trajectories necessary to reach all points in the conservative approximation of the target set.
The method demonstrates optimality under active constraints as $\Yt(T_f) \subset \mathcal{P}^3 \subseteq \X_{\delta}(T_f)$, $\partial \mathcal{P}^3 \cap \partial \X_{\delta}(T_f) \neq \emptyset$ and  $\x_{nom}(T_f) \in \Yt(T_f)$.
\section{Conclusions}\label{section:Conclusions}
In this paper, we have developed and implemented a novel method for reachability analysis in $SE(3)$ and control under uncertainty, applying it to the interception of dynamic targets in a space scenario. 
Our \acrlong{RG-OCP} formulation ensures robust feasibility and guarantees reachability to dynamic targets subject to uncertainty, addressing a significant challenge in modern robotics. 
First, we have proposed a conservative method to estimate reachable sets on $SE(3)$, which we have then embedded within an \gls{OCP} to formulate the \gls{RG-OCP}.
We have validated the effectiveness of the proposed method in generating feasible and optimal interception trajectories within the constraints of bounded uncertainty.
The nonlinearity in the problem slows its resolution and currently limits online applicability. 
Moreover, tighter theoretical results could be obtained to describe reachable sets on $SO(3)$.
Therefore, future work will address the introduction of parallel computation through GPU acceleration, foundational work to obtain tighter theoretical guarantees and further applications to real-world scenarios, addressing parametric and system uncertainty. 
\section{Aknowledgements}
The authors kindly thank Riccardo Bonalli for his feedback and insight.
\printbibliography

@misc{lew_estimating_2023,
	title = {Estimating the Convex Hull of the Image of a Set with Smooth Boundary: Error Bounds and Applications},
	number = {{arXiv}:2302.13970},
	publisher = {{arXiv}},
	author = {Lew, Thomas and Bonalli, Riccardo and Janson, Lucas and Pavone, Marco},
	urldate = {2024-03-01},
	date = {2023-07-10},
}

@misc{lew_convex_2024,
	title = {Convex Hulls of Reachable Sets},
	number = {{arXiv}:2303.17674},
	publisher = {{arXiv}},
	author = {Lew, Thomas and Bonalli, Riccardo and Pavone, Marco},
	date = {2024-02-29},
}

@inproceedings{deits2015computing,
  title={Computing large convex regions of obstacle-free space through semidefinite programming},
  author={Deits, Robin and Tedrake, Russ},
  booktitle={Algorithmic Foundations of Robotics XI: Selected Contributions of the Eleventh International Workshop on the Algorithmic Foundations of Robotics},
  year={2015},
  organization={Springer}
}

@article{liu2017planning,
  title={Planning dynamically feasible trajectories for quadrotors using safe flight corridors in 3-d complex environments},
  author={Liu, Sikang and Watterson, Michael and Mohta, Kartik and Sun, Ke and Bhattacharya, Subhrajit and Taylor, Camillo J and Kumar, Vijay},
  journal={IEEE Robotics and Automation Letters},
  year={2017},
  publisher={IEEE}
}

@book{do1992riemannian,
  title={Riemannian geometry},
  author={Do Carmo, Manfredo Perdigao and Flaherty Francis, J},
  volume={2},
  year={1992},
  publisher={Springer}
}

@article{gudmundsson2004introduction,
  title={An introduction to Riemannian geometry},
  author={Gudmundsson, Sigmundur},
  journal={Lecture Notes version},
  volume={1},
  number={9},
  year={2004}
}

@article{hartley2010rotation,
  title={Rotation averaging and weak convexity},
  author={Hartley, Richard and Trumpf, Jochen and Dai, Yuchao and others},
  year={2010},
  publisher={Institute of Electrical and Electronics Engineers (IEEE Inc)}
}

@article{albee2021robust,
  title={A robust observation, planning, and control pipeline for autonomous rendezvous with tumbling targets},
  author={Albee, Keenan and Oestreich, Charles and Specht, Caroline and Ter{\'a}n Espinoza, Antonio and Todd, Jessica and Hokaj, Ian and Lampariello, Roberto and Linares, Richard},
  journal={Frontiers in Robotics and AI},
  volume={8},
  pages={641338},
  year={2021},
  publisher={Frontiers Media SA}
}

@inproceedings{hillenbrand2005motion,
  title={Motion and parameter estimation of a free-floating space object from range data for motion prediction},
  author={Hillenbrand, Ulrich and Lampariello, Roberto},
  booktitle={Proceedings of i-SAIRAS},
  year={2005}
}

@article{lampariello2021robust,
  title={Robust motion prediction of a free-tumbling satellite with on-ground experimental validation},
  author={Lampariello, Roberto and Mishra, Hrishik and Oumer, Nassir W and Peters, Jan},
  journal={Journal of Guidance, Control, and Dynamics},
  volume={44},
  number={10},
  pages={1777--1793},
  year={2021},
  publisher={American Institute of Aeronautics and Astronautics}
}

@inproceedings{smith2016astrobee,
  title={Astrobee: A new platform for free-flying robotics on the international space station},
  author={Smith, Trey and Barlow, Jonathan and Bualat, Maria and Fong, Terrence and Provencher, Christopher and Sanchez, Hugo and Smith, Ernest},
  booktitle={i-SAIRAS},
  number={ARC-E-DAA-TN31584},
  year={2016}
}

@article{Tordesillas_2022,
  author       = {Jesus Tordesillas and
                  Jonathan P. How},
  title        = {{PANTHER:} Perception-Aware Trajectory Planner in Dynamic Environments},
  journal      = {CoRR},
  year         = {2021},
}

@misc{gao_closure_2024,
	title = {{CLOSURE}: Fast Quantification of Pose Uncertainty Sets},
	publisher = {{arXiv}},
	author = {Gao, Yihuai and Tang, Yukai and Qi, Han and Yang, Heng},
	date = {2024-03-14},
}

@inproceedings{lew_exact_2023,
	location = {Singapore, Singapore},
	title = {Exact Characterization of the Convex Hulls of Reachable Sets},
	booktitle = {2023 62nd {IEEE} Conference on Decision and Control (CDC)},
	publisher = {{IEEE}},
	author = {Lew, Thomas and Bonalli, Riccardo and Pavone, Marco},
	date = {2023-12-13},
}

@article{lathrop2021distributionally,
  title={Distributionally Safe Path Planning: Wasserstein Safe {RRT}},
  author={Lathrop, Paul and Boardman, Beth and Mart{\'i}nez, Sonia},
  journal={Robotics and Automation Letters},
  volume={7},
  number={1},
  pages={430--437},
  year={2021},
}

@article{lindemann2021robust,
  author       = {Lars Lindemann and
                  Matthew Cleaveland and
                  Yiannis Kantaros and
                  George J. Pappas},
  title        = {Robust Motion Planning in the Presence of Estimation Uncertainty},
  journal      = {CoRR},
  year         = {2021},
}

@article{abbasi2011improved,
  title={Improved algorithms for linear stochastic bandits},
  author={Abbasi-Yadkori, Yasin and P{\'a}l, D{\'a}vid and Szepesv{\'a}ri, Csaba},
  journal={Advances in neural information processing systems},
  volume={24},
  year={2011}
}

@inproceedings{singh2020robust,
  title={Robust tracking with model mismatch for fast and safe planning: an sos optimization approach},
  author={Singh, Sumeet and Chen, Mo and Herbert, Sylvia L and Tomlin, Claire J and Pavone, Marco},
  journal      = {CoRR},
  year={2018},
  organization={Springer}
}

@ARTICLE{danielson2020robust,  
    author={C. {Danielson} and K. {Berntorp} and A. {Weiss} and S. D. {Cairano}},  
    journal={IEEE Transactions on Automatic Control},   
    title={Robust Motion Planning for Uncertain Systems With Disturbances Using the Invariant-Set Motion Planner},   
    year={2020},  
}

@article{majumdar2016,
author = {Majumdar, Anirudha and Tedrake, Russ},
year = {2016},
title = {Funnel Libraries for Real-Time Robust Feedback Motion Planning},
volume = {36},
number = {8},
journal = {The Int. Journal of Robotics Research},
}

@Inproceedings{lew2020samplingbased,
    title        = {Sampling-based Reachability Analysis: A Random Set Theory Approach with Adversarial Sampling},
    author       = {Lew, T. and Pavone, M.},
    booktitle    = {Conference on Robot Learning},
    year         = {2020},
	asl_abstract = {Reachability analysis is at the core of many applications, from neural network verification, to safe trajectory planning of uncertain systems. However, this problem is notoriously challenging, and current approaches tend to be either too restrictive, too slow, too conservative, or approximate and therefore lack guarantees. In this paper, we propose a simple yet effective sampling-based approach to perform reachability analysis for arbitrary dynamical systems. Our key novel idea consists of using random set theory to give a rigorous interpretation of our method, and prove that it returns sets which are guaranteed to converge to the convex hull of the true reachable sets. Additionally, we leverage recent work on robust deep learning and propose a new adversarial sampling approach to robustify our algorithm and accelerate its convergence. We demonstrate that our method is faster and less conservative than prior work, present results for approximate reachability analysis of neural networks and robust trajectory optimization of high-dimensional uncertain nonlinear systems, and discuss future applications.},
	asl_month    = aug,
  	asl_url      = {https://arxiv.org/abs/2008.10180},
}

@Inproceedings{LewJansonEtAl2022,
author = {Lew, T. and Janson, L. and Bonalli, R. and Pavone, M.},
title = {A Simple and Efficient Sampling-based Algorithm for General Reachability Analysis},
year = {2022},
booktitle = {Learning for Dynamics \& Control Conference},
}

@INPROCEEDINGS{ZhangTargetInterception,
  author={Zhang, Xuebo and Wang, Yongxin and Fang, Yongchun},
  booktitle={2016 IEEE International Conference on Real-time Computing and Robotics (RCAR)}, 
  title={Vision-based moving target interception with a mobile robot based on motion prediction and online planning}, 
  year={2016},
 }

@ARTICLE{DongkyoungInterceptionRobotArm,
  author={Dongkyoung Chwa and Junho Kang and Jin Young Choi},
  journal={IEEE Transactions on Systems, Man, and Cybernetics}, 
  title={Online trajectory planning of robot arms for interception of fast maneuvering object under torque and velocity constraints}, 
  year={2005}
  }

@article{mark2019review,
  title={Review of active space debris removal methods},
  author={Mark, C Priyant and Kamath, Surekha},
  journal={Space policy},
  volume={47},
  pages={194--206},
  year={2019},
  publisher={Elsevier}
}

@book{Lee2012,
  title 				   = {Introduction to Smooth Manifolds},
  author 				   = {Lee, J.M.},
  year 					   = {2012},
  publisher 			   = pub_Springer_NY,
  edition   = {Second},
}

@book{Lee2018,
  title 				   = {Introduction to {Riemannian} Manifolds},
  author 				   = {Lee, J.M.},
  year 					   = {2018},
  publisher 			   = pub_Springer,
  edition   = {Second},
}

\appendices{
\section*{Appendix: Proof of Theorem \ref{Theo1}}
\textit{Proof:} From \cite{lew_exact_2023}, it holds true that 
\begin{equation}
    \Xi(t) \subseteq  H(\Xi_{\delta}(t)) \oplus B(0, \varepsilon)
    \label{eq:LewTheorem}
\end{equation} 
at all times $t$. 
By definition of the \gls{RTC} (\ref{eq:TimeBoundEllipsoid}), $\x_k(t) \in \mathcal{E}_{k,i}$ for any $ t_i \leq t \leq t_{i+1}$. 
Thus
\begin{equation}
    \Xi_{\delta}(t) \subset \bigcup_k^M \mathcal{E}_{k,i}.
\end{equation}
The condition holds at all times $t$: 
\begin{equation}
    \bigcup_{t=t_i}^{t_{i+1}} \Xi_{\delta}(t) \subset \bigcup_k^M \mathcal{E}_{k,i}.
\end{equation}
Now, applying the union operator to (\ref{eq:LewTheorem}): 
\begin{align}
    \bigcup_{t=t_i}^{t_{i+1}}\Xi(t) &\subseteq \bigcup_{t=t_i}^{t_{i+1}} \big[H(\Xi_{\delta}(t)) \oplus B(0, \varepsilon)\big] \nonumber \\
    &\subset \bigcup_{t=t_i}^{t_{i+1}} \big[H(\bigcup_k^M \mathcal{E}_{k,i}) \oplus B(0, \varepsilon)\big] \nonumber 
\end{align}
by the properties of the Minkowski sum, we can write: 
\begin{equation}
    H(\bigcup_k^M \mathcal{E}_{k,i}) \oplus B(0, \varepsilon) = H(\bigcup_k^M \mathcal{E}_{k,i} \oplus B(0, \varepsilon)) = \mathcal{R}_i \nonumber
\end{equation}
thus obtaining: 
\begin{equation}
    \bigcup_{t=t_i}^{t_{i+1}}\Xi(t) \subset \bigcup_{t=t_i}^{t_{i+1}} \mathcal{R}_i = \mathcal{R}_i
\end{equation}
as $\mathcal{R}_i$ is constant for time $t_i \leq t \leq t_{i+1}$.
This results concludes the proof, demonstrating that the \gls{RTC} completely encloses all reachable sets in a time-interval.
}

\end{document}